# SmoothSegNet: A Global-Local Framework for Liver Tumor Segmentation with Clinical Knowledge-Informed Label Smoothing

Hairong Wang*, Lingchao Mao*, Zihan Zhang, Jing Li

H. Milton Stewart School of Industrial and Systems Engineering, Georgia Institute of Technology, Atlanta, GA, USA

*Contributed equally to the manuscript

Corresponding Author: Jing Li (Email: jing.li@isye.gatech.edu)

## Abstract

Liver cancer is a leading cause of mortality worldwide, and accurate Computed Tomography (CT)-based tumor segmentation is essential for diagnosis and treatment. Manual delineation is time-intensive, prone to variability, and highlights the need for reliable automation. While deep learning has shown promise for automated liver segmentation, precise liver tumor segmentation remains challenging due to the heterogeneous nature of tumors, imprecise tumor margins, and limited labeled data. We present SmoothSegNet, a novel deep learning framework that addresses these challenges with the three key designs: (1) A novel knowledge-informed label smoothing technique that distills knowledge from clinical data to generate smooth labels, which are used to regularize model training, reducing the overfitting risk and enhancing model performance; (2) A global and local segmentation framework that breaks down the main task into two simpler sub-tasks, allowing optimized preprocessing and training for each; and (3) Pre- and post-processing pipelines customized to the challenges of each subtask aimed to enhance tumor visibility and refines tumor boundaries. We apply the proposed model on a challenging HCC-TACE-Seg dataset and show that SmoothSegNet outperformed various benchmarks in segmentation performance, particularly at smaller tumors (<10cm). Our ablation studies show that the three design components complementarily contribute to the model improved performance. Code for the proposed method are available at https://github.com/lingchm/medassist-liver-cancer.

## 1. Introduction

Liver cancer is a significant global health concern, affecting over 800,000 people annually and standing among the leading causes of cancer-related deaths worldwide (American Cancer Society, 2024). Liver is also a common destination for metastatic cancer cells originating from various abdominal organs, including the colon, rectum, pancreas, as well as distant organs such as the breast and lung. Consequently, a thorough examination of the liver and its lesions is critical to comprehensive tumor staging and management strategies. Standard tumor assessment protocols require precise measurement of the diameter of the largest target lesion (Eisenhauer et al., 2009). Thus, accurate localization and precise segmentation of liver tumors within Computed Tomography (CT) scans are essential for effective diagnosis, treatment planning, and monitoring of therapeutic response in patients with liver cancer (Terranova & Venkatakrishnan, 2024; Virdis et al., 2019). Manual delineation of target lesions in CT scans is fraught with challenges, being both time-consuming and prone to poor reproducibility and operator-dependent variability (Gul et al., 2022). Automated liver tumor segmentation can provide rapid and consistent tumor delineation, thereby improving patient outcomes and reducing healthcare costs.

Although many deep learning (DL) algorithms achieved promising performance in automated liver segmentation recently, with dice scores ranging from 0.90 to 0.96, enhancing automated liver tumor segmentation remains a challenge, currently standing at dice scores from 0.41 to 0.67 according to a recent Liver Tumor Segmentation Benchmark . Liver tumor segmentation is an inherently challenging task because tumors vary significantly in size, shape, and location across different patients, which leads to a broad spectrum of tumor characteristics and hinders model generalization (Sabir et al., 2022). Moreover, margins of some tumors are imprecise as CT scans exhibit low gentile brightness and roughness making it difficult to distinguish between tumorous and healthy liver tissue (Sabir et al., 2022). The scarcity of labeled data, coupled with the variations in lesion-to-background contrast, the coexistence of different lesion types, and disease-specific variability introduces substantial challenges in developing deep learning algorithms for accurate liver tumor segmentation (Moghbel et al., 2018).

Various studies have emerged to tackle the challenges of tumor segmentation. Recent advancements have primarily focused on optimizing network architecture design to enhance segmentation performance. Some researchers have integrated components like a variational auto-encoder (VAE) branch to regularize shared layers, thereby stabilizing the training process and improving generalization (Myronenko, 2018). Others have explored more complex architectures such as combining ResNet and U-Net models, leveraging ResNet's deep feature extraction capabilities with U-Net's segmentation proficiency, achieving gains in quality and computational efficiency (Rahman et al., 2022). Similarly, transformers have been employed to capture features across varying Magnetic Resonance Imaging (MRI) contrast resolutions, enhancing segmentation accuracy through self-attention mechanisms (Hatamizadeh et al., 2022). Traditional machine learning methods such as multi support vector machines (MSVMs) have also been applied to tumor segmentation, typically by classifying handcrafted features extracted from image patches (Hasuni Shahrbabak et al., 2022). Although these models achieve near-perfect accuracy in liver segmentation, their tumor segmentation performance remains poor due to the low liver-to-tumor contrast and irregular morphology of tumors. We hypothesize that liver and tumor segmentation are two tasks distinct enough that require different feature representations and signals for accurate segmentation. Hence, we explore a two-step strategy that independently optimizes liver and tumor segmentations. By dividing and conquering liver and tumor segmentation, our method captures the unique characteristics of each while offering greater flexibility in refining the training as well as pre- and post-processing of each task.

To address the challenge of limited labeled data in medical image segmentation, researchers have adopted methods that incorporate additional information to supervise the models. For instance, tumor bounding boxes from unlabeled data have been used to guide the student module in a teacher-student learning framework, improving model performance with minimal manual annotation (D. Zhang et al., 2021). Other works have leveraged software-generated pseudo-labels, allowing models to learn from large datasets with reduced human intervention (Lyu et al., 2022). Fine-tuning models using a combination of unlabeled data and generated pseudo-labels has also been shown to yield continuous improvements in segmentation accuracy (Chen et al., 2024). Some studies have focused on pooling data from multiple sites to enhance model generalization through transfer learning and domain adaptation (You et al., 2022). We propose to leverage clinical data to help with learning the segmentation task, as several clinical risk factors such as excessive alcohol consumption and smoking, are known to significantly influence liver cancer. The insights from these risk factors are a form of biomedical knowledge that can be used to supervise model training.

Integrating biomedical domain knowledge has been a promising avenue to alleviate data shortages and improve model performance in training DL models (Mao et al., 2024). For instance, some researchers have developed customized architectures based on biological knowledge (Fortelny & Bock, 2020). Mathematical oncology models, such as tumor growth models, have been used to regularize loss functions (Gaw et al., 2019; L. Wang et al., 2022) or to formulate Physics Informed Neural Networks (PINN) (Meaney et al., 2023). When knowledge of certain regions of images that are most relevant to the prediction is available, researchers employed attention mechanisms to guide the model in focusing on important areas (Tomita et al., 2019). Other researchers have proposed incorporating feature behavior knowledge as attribution priors during training, such as hierarchical relationships among different labels (H. Wang et al., 2024) or ordinal relationships among unlabeled samples (Mao et al., 2023; L. Wang et al., 2024). However, all these existing methods require explicit domain knowledge for model design. We propose a novel strategy for distilling knowledge from the clinical data and leveraging it to generate soft labels to regularize model training. Unlike the conventional approach of incorporating clinical data as additional predictors, our method does not require clinical data at inference time, enabling the model to make predictions for new patients when their clinical data is unavailable.

In this work, we propose a deep learning framework for automated liver and tumor segmentation called SmoothSegNet. This framework is designed to address the above-mentioned challenges. The main contributions of this work are summarized in Table 1 and outlined as follows:

- **A novel knowledge-informed label smoothing technique**: We distill knowledge from clinical data to derive an imprecise tumor size label through an interpretable regression model, which is then used to regularize model training. This imprecise tumor size label encodes auxiliary medical knowledge, helping to mitigate overfitting in small-sample settings and enhance model generalization.

- **A global and local framework optimized for liver and tumor segmentation sub-tasks**: We propose a sequential framework that begins with a global-view liver segmentation model, followed by a local-view tumor segmentation model. This two-step approach offers greater flexibility compared to the traditional multi-class approach, enabling the preprocessing, training, and post-processing to be optimized for each subtask. We incorporate an active contour algorithm for refinement of the segmented tumor boundaries, improving the segmentation results for tumors with imprecise margins.
- **Contribution to liver cancer diagnosis and treatment:** We apply SmoothSegNet to a real-data application for segmenting liver tumor from CT scans of patients with Hepatocellular carcinoma. SmoothSegNet generated more accurate segmentations compared to various state-of-the-art deep learning models, particularly improving the performance for small tumors.

Table 1. A summary presents research gaps, existing methods, and the key contributions.

| | Research Gap | Existing Methods | Contributions of Proposed Method |
|---|---|---|---|
| **Clinical Challenge** | Liver tumor segmentation inaccurate due to tumor heterogeneity | Gul et al., 2022; Bilic et al., 2023; Sabir et al., 2022; Moghbel et al., 2018 | Provide a general framework for liver tumor segmentation with significantly improved performance |
| **Technical Challenge** | Lack of domain knowledge | D. Zhang et al., 2021; Lyu et al., 2022; S. Chen et al., 2024; You et al., 2022; Mao et al., 2024; Fortelny & Bock, 2020; Gaw et al., 2019; L. Wang et al., 2022; Meaney et al., 2023; Tomita et al., 2019; H. Wang et | Provide a solution for effectively distilling knowledge from clinical data and leveraging it to guide model training while preventing overfitting; |

| | | al., 2024; Mao et al., 2023; L. Wang et al., 2024 | Able to make predictions for new patients when clinical data is unavailable |
|---|---|---|---|
| | Different characteristics of organs and tumors overlooked | Myronenko, 2018; Rahman et al., 2022; Hatamizadeh et al., 2022 | Exploit the characteristics of livers and tumors separately; Provide flexibility in leveraging pre- and post-processing for segmentation masks |
| **Data Challenge** | Data scarcity | Moghbel et al., 2018; Khalaf et al., 2018; Moawad et al., 2021 | Provide state-of-the-art results for HCC tumor segmentation |

The remainder of this paper is organized as follows. Section 2 describes the details of the proposed framework, structured around the three key contributions. Section 3 presents a case study evaluating the effectiveness of the proposed model. Section 4 concludes with a summary of findings and future directions.

## 2. SmoothSegNet: Global-Local Framework with Knowledge-Informed Label Smoothing

The proposed segmentation framework consists of three building blocks: **knowledge-distillation from clinical data**, **liver segmentation** using *global-view* inputs, and **tumor segmentation** using *local-view* inputs, as shown in Figure 1. Each segmentation block involves its own design for input preprocessing, model training, and post-processing of segmentation outputs optimized for either liver or tumor segmentation sub-task. In the following subsections, we detail the proposed knowledge-informed label smoothing technique, and the global and local-view segmentation framework, respectively.

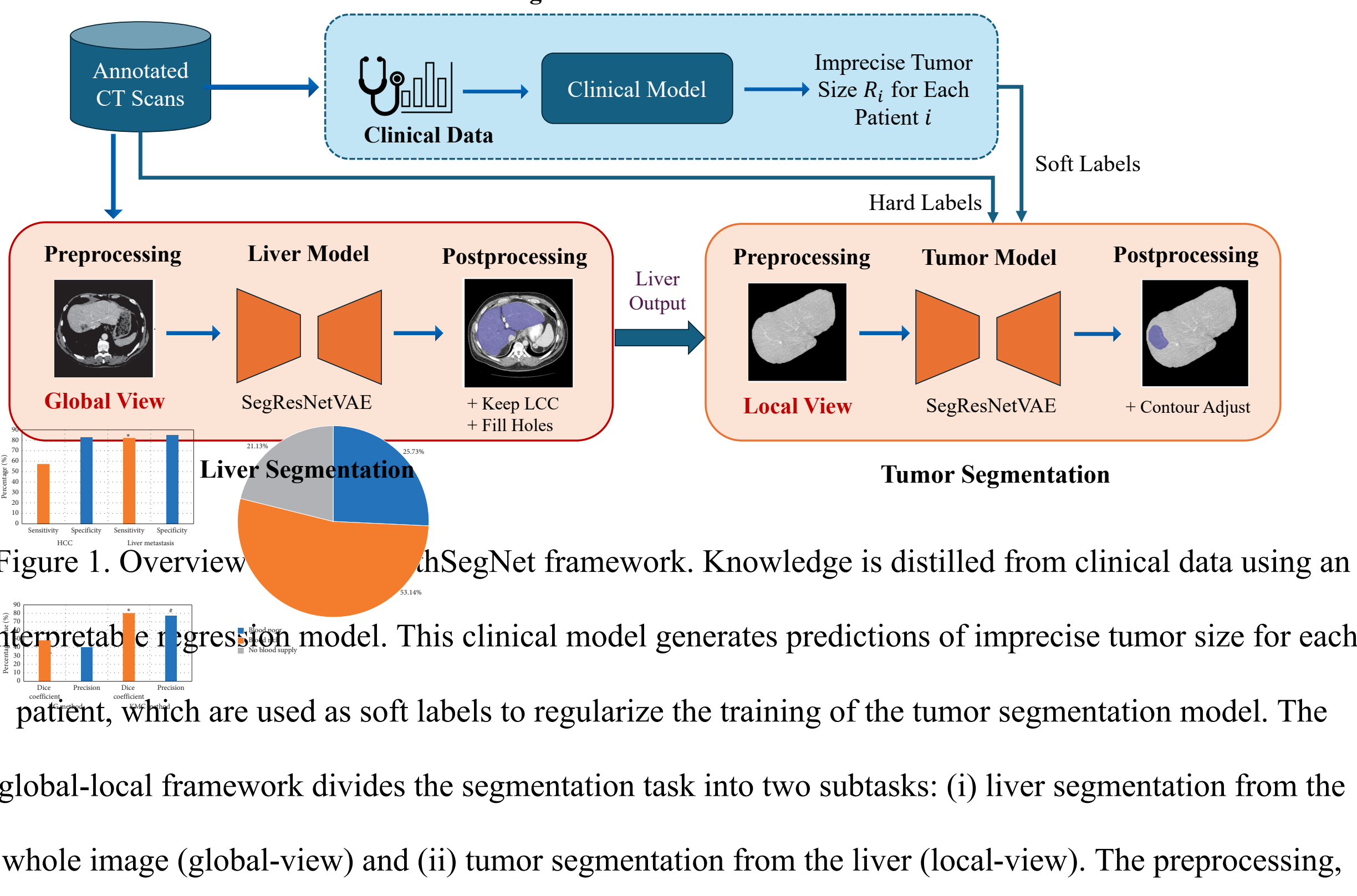

Figure 1. Overview [illegible]hSegNet framework. Knowledge is distilled from clinical data using an interpretable regression model. This clinical model generates predictions of imprecise tumor size for each patient, which are used as soft labels to regularize the training of the tumor segmentation model. The global-local framework divides the segmentation task into two subtasks: (i) liver segmentation from the whole image (global-view) and (ii) tumor segmentation from the liver (local-view). The preprocessing, post-processing, and training details are optimized for each sub-task.

### 2.1 Clinical Knowledge-Informed Label Smoothing

The significant heterogeneity of tumor shapes, sizes, and intensities across patients is widely observed in liver tumors. A typical training approach that minimizes segmentation loss on a sample-by-sample basis can easily cause overfitting. It often causes the model to become overly confident in its predictions, especially when the dataset is limited and contains label noise. Label smoothing has become a widely used regularization strategy in many state-of-the-art models. It injects noise by softening the target labels, which prevents the model from becoming overconfident in its predictions. This forces the model to consider all classes with some probability, improving generalization and reducing the risk of overfitting. Various empirical evidence showed that label smoothing effectively prevents networks from becoming over-confident and encourages the representations of training examples from the same class to form tight clusters (Müller et al., 2020; Szegedy et al., 2015; C.-B. Zhang et al., 2021). By using soft targets that are a weighted average of the hard labels and the uniform distribution over labels, the generalization of a neural network can be improved (C.-B. Zhang et al., 2021). However, most existing label smoothing techniques are typically designed for classification problems, ignoring underlying properties of dense prediction problems, such as medical image segmentation (Vasudeva et al., 2024).

Building on the principles of traditional label smoothing, we introduce a novel knowledge-informed label smoothing technique that distills knowledge from an external clinical dataset to generate soft labels. These soft labels serve as imprecise labels of tumor size, incorporating clinically derived insights to enhance the segmentation. More specifically, an interpretable regression model is trained to estimate the rough volume size of tumor through clinically validated risk factors for liver cancer. Then, the imprecise tumor size label is used to regularize the training of the tumor segmentation model, facilitating noise injection and preventing the model from becoming overconfident. This approach strikes a balance between noise injection and medical knowledge guidance for model training. The knowledge-informed label smoothing process is illustrated in Figure 2. From a model design perspective, instead of directly using risk factors as external variables, the imprecise tumor size label guides the model behavior to align closely with clinical expectations during training and does not require the availability of such clinical data at inference time. From an empirical perspective, our experimental results show that the proposed knowledge-informed label smoothing technique is effective in preventing overfitting and improving generalization.

The following subsections detail our approach to knowledge distillation from clinical data and its integration as soft labels in the training process.

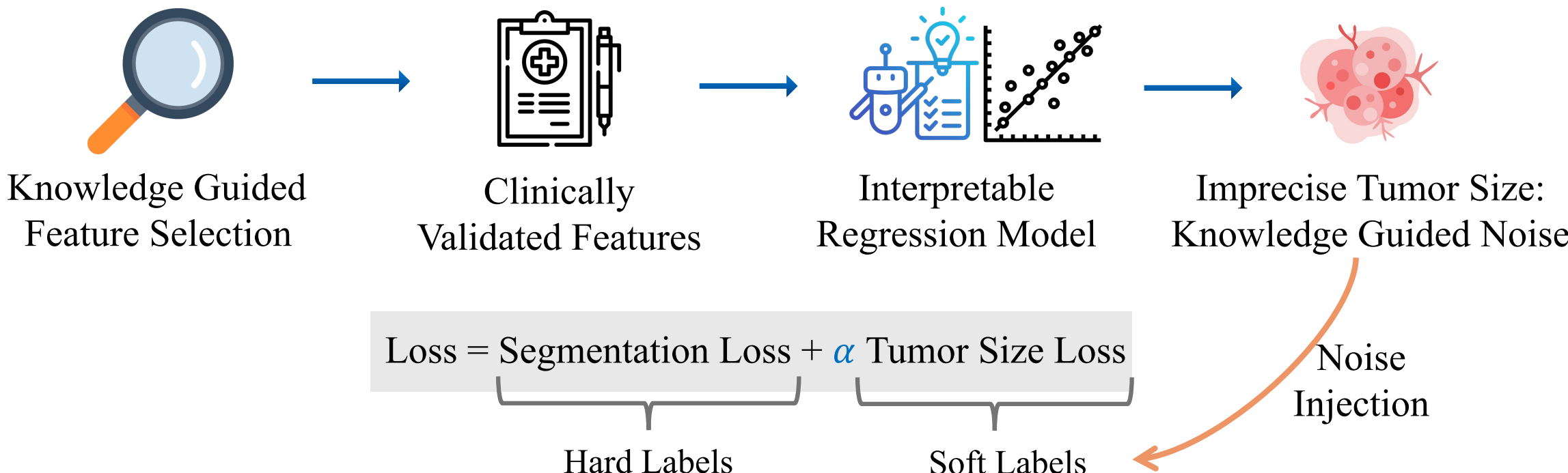

Figure 2. The knowledge-informed label smoothing process, where $\alpha$ is a hyper-parameter to weight the influence of soft labels.

### 2.1.1 Knowledge Distillation from Clinical Data

Knowledge distillation (KD), originally introduced for model compression, is a versatile technique that enables training a model on an auxiliary dataset while transferring the learned knowledge to the main model. For instance, KD has been used to train a teacher model on the data from one modality, which then supervises a student model learning from a different modality (Kwak et al., 2023). KD can also be applied within the same dataset; for example, self-knowledge distillation has been employed to enforce consistency between predictions on augmented and original samples, serving as a regularization strategy to mitigate overfitting (Xu & Liu, 2019).

In this work, we propose leveraging knowledge distillation to transfer the knowledge from clinical dataset to assist with the segmentation task. Specifically, we develop a model that captures the relationship between clinical risk factors and tumor size across all the patients. This model distills knowledge from clinical data and produces imprecise tumor size labels, which are subsequently used as soft labels to guide segmentation model training. By incorporating these clinically informed soft labels, we aim to enhance the model's ability to generalize across diverse tumor presentations, improving robustness and accuracy in liver tumor segmentation.

We develop an interpretable regression model based on relevant risk factors from the clinical dataset to generate imprecise tumor size labels across all the patients. Building on existing research regarding the risk factors for liver cancer (Mohammadian et al., 2018; Morshid et al., 2019), we began by identifying the relevant risk factors present in the clinical dataset that are known to potentially influence the progression of liver cancer. Several well-studied risk factors significantly influence disease progression, including chronic infections with hepatitis B (HBV) and hepatitis C (HCV) viruses, excessive alcohol consumption, smoking, metabolic conditions such as obesity and diabetes, and a family history of liver cancer (El-Serag & Mason, 2000; Mohammadian et al., 2018). These factors not only drive the initiation and progression of tumors but also contribute to its heterogeneity. On the other hand, clinical prognostic measurements such as the Cancer of the Liver Italian Program (CLIP) score and Alpha-fetoprotein (AFP) levels can be informative indicators of disease severity (Morshid et al., 2019). Incorporating insights from these risk factors to guide the training of tumor segmentation models can enhance model accuracy and reliability.

Specifically, we define the Tumor-to-Liver Volume Ratio (TLVR) to quantify the overall tumor size relative to the liver size, denoted as $R_i$ for patient $i$ as:

$$R_i = \frac{number\ of\ voxels\ in\ tumor\ mask_i}{number\ of\ voxels\ in\ liver\ mask\ including\ tumor_i} \tag{1}$$

We trained an interpretable regression models with TLVR extracted from segmentation masks as the dependent variable and the clinical risk factors as independent variables. This model synthesizes clinical insights about the tumor size for this patient cohort. This intrinsic relationship established through this knowledge extraction stage is then used to regularize model training, as described in the next section.

#### 2.1.2 Learning with Knowledge-Informed Smooth Labels

Liver tumor segmentation is a classic challenging task due to the inherent noise in the dataset. This noise arises from the wide variability in tumor shapes, as well as blurry and irregular boundaries, leading to poor inter-observer agreement even among experts (Gul et al., 2022). Training deep learning models on a small size dataset of this nature often results in overfitting, severely compromising generalization and leading to high false positive and false negative rates in segmentation outputs. To address this issue, we draw inspiration from label smoothing and propose a novel approach that leverages knowledge distilled from clinical data to generate smooth labels. Instead of adding a non-informative uniform distribution over labels, our knowledge-informed smooth labels provide medical guidance to the segmentation task.

Specifically, the clinical model described in Section 2.1.1 is used to generate predictions of TLVR, which serves as clinically informed soft labels for each patient. This soft label provides an image-level, imprecise representation of tumor size, while the ground truth tumor segmentation mask offers pixel-level, precise measurement of tumor presence. The tumor segmentation model is optimized by minimizing the loss on both the original hard labels and knowledge-informed soft labels:

$$L_{tumor} = L_{hard} + \alpha L_{soft}, \tag{2}$$

where $\alpha$ is a tunable hyper-parameter to weight the influence of soft labels.

We use the Mean Squared Loss to encourage consistency of the predicted segmentation mask with the knowledge-informed soft label:

$$L_{soft} = \frac{1}{n}\sum_{i}\left(R_i - \hat{R}_i\right)^2 . \tag{3}$$

We use the standard dice and focal loss to learn from the segmentation mask:

$$L_{hard} = \lambda_f L_{focal} + \lambda_d L_{dice}, \tag{4}$$

where $\lambda_f, \lambda_d$ are tunable hyper-parameters to weight the importance of each loss. $L_{dice}$ is dice loss (Hänsch et al., 2022) commonly used for segmentation models,

$$L_{dice} = 1 - \frac{2\sum_{i=1}^{N} p_i g_i}{\sum_{i=1}^{N} p_i^2 + \sum_{i=1}^{N} g_i^2} , \tag{5}$$

where the sums run over the $N$ voxels, of the predicted segmentation volume $p_i \in P$ and the ground truth binary volume $g_i \in G$. $L_{focal}$ is focal loss (Lin et al., 2017) that accounts for imbalanced segmentation,

$$L_{focal} = -\sum_{i=1}^{N} (i - p_i)^{\gamma} \log(p_i) , \tag{6}$$

where $\gamma \geq 0$ is the focusing parameter.

By training the model using both hard labels and soft labels, SmoothSegNet can take into consideration high-level clinical trends across patients while optimizing on images of individual patients. This dual-objective regularizes the learning process, helping the model better generalize and improving its robustness to data noise and variability. Our empirical results show that the label smoothing technique effectively reduced overfitting during model training (Appendix D). It is worth noting that another advantage of leveraging clinical data through knowledge distillation, rather than training multimodal models that use clinical data as additional features, is that it eliminates the need for clinical data at inference time. This ensures that the model can still make predictions for new patients, even when their clinical data is unavailable.

### 2.2 A Global and Local-view Segmentation Framework

While knowledge-informed label smoothing helps reduce overfitting, we adopt another strategy to improve segmentation performance. Liver tumor segmentation remains a challenging task because the lesion comprises a mix of small and large tumors with diverse shapes and boundary characteristics (Bilic et al., 2023; Hänsch et al., 2022). Our hypothesis that segmenting the liver from the abdomen and segmenting the tumor from the liver are two distinct tasks, each presenting unique modeling challenges and requiring different sources of informative signals. As such, using a single model for both tasks could constraint its effectiveness. Hence, we propose a two-step framework breaking down the original task of "*segmenting liver tumors from abdomen*" into one simple auxiliary sub-task and one primary sub-task: (i) "*segmenting liver from abdomen*" and (ii) "*segmenting tumor from liver.*"

In the first sub-task, a segmentation model focuses on delineating the liver from other structures within the whole abdominal CT image (*global view*). This task requires comprehensive understanding of anatomical structures. Since the location and shape of livers compared to other organs is generally consistent across patients, existing deep learning models have solved this problem with high accuracy. Utilizing the liver masks obtained from the first task, the CT image is cropped to a liver-focused bounding box from which a second model segments the tumor (*local view*). The model's focus narrows to the liver area, demanding the capture of precise, fine-grained feature information to distinguish between healthy and tumorous tissues within the organ. Unlike liver segmentation where every liver's location and shapes are relatively similar, tumor segmentation presents unique challenges of tumor heterogeneity in number, sizes, shapes, and intensities. We tackle this challenge by introducing a knowledge-informed label smoothing technique, as described in Section 2.1. Additionally, some tumors have imprecise boundaries that are not easily distinguishable. To tackle this issue, we employed an active contour adjustment algorithm to refine tumor contours, as detailed in Appendix C and Supplementary C.

The sequential two-model design offers greater flexibility compared to traditional multi-class methods, as it enables preprocessing, model training, hyper-parameter tuning, and post-processing to be optimized for the performance of each subtask. Appendix C and Supplementary C provides a sample pre- and post-processing pipeline tailed for liver and tumor segmentation tasks that we used in this study.

## 3. Application in Hepatocellular Carcinoma

Hepatocellular carcinoma (HCC) is the most common liver cancer, with rising incidence and limited treatment options like transarterial chemoembolization (TACE), which fails in up to 60% of cases (S. Zhang et al., 2022). We used the HCC-TACE-Seg dataset from the Cancer Imaging Archive (TCIA), which includes CT scans of 105 HCC patients treated with Transarterial Chemoembolization (TACE) (Moawad et al., 2021). Further dataset details are provided in Supplementary A.

### 3.1 Clinical Indicator of TLVR

Based on the experimental results, linear regression demonstrated relatively strong performance in predicting TLVR among various linear and non-linear models. Due to its interpretability, we selected linear regression for further analysis. The outcomes of the model, including the significance of each feature, are shown in Figure 3. The Pearson correlation coefficient between $R$ and $\hat{R}$ is 0.820, indicating a strong correlation and highlighting the predictive power of the clinical features in relation to TLVR. To minimize the risk of overfitting and enhance model interpretability, we selected the features with the highest contributions for TLVR prediction. The detailed feature selection process is outlined in Appendix A.

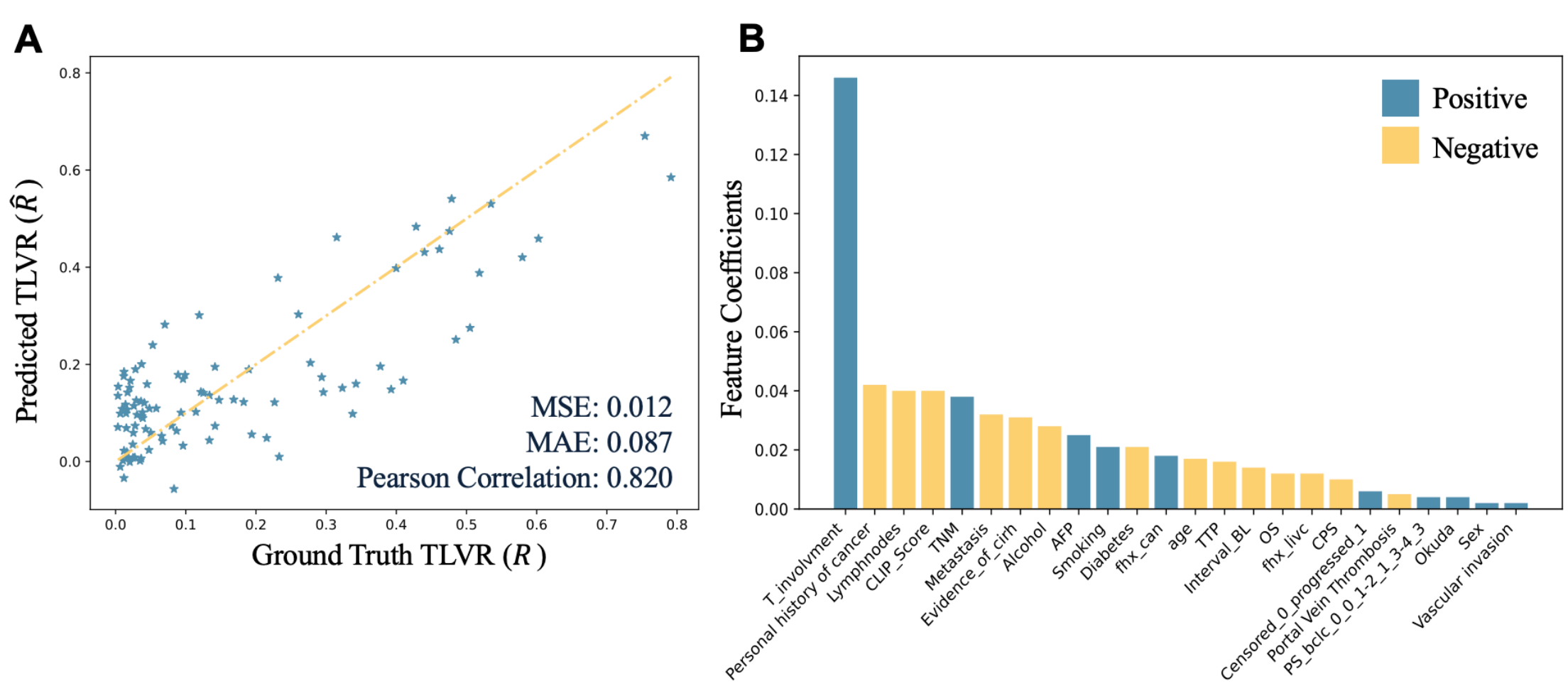

Figure 3. (A) Results of the linear regression, showcasing the relationship between clinical features and the tumor-to-liver volume ratio (TLVR). (B) The coefficient plot indicating feature contributions. MSE: mean squared error; MAE: Mean Absolute Error.

### 3.2 Segmentation Results

We compared SmoothSegNet against several state-of-the-art medical image segmentation models, including U-Net (Kerfoot et al., 2019), U-Net++ (Z. Zhou et al., 2018), SegResNet (Myronenko, 2018), SegResNetVAE (Myronenko, 2018), and MA-Net (Fan et al., 2020). Details of the benchmark methods and evaluation metrics can be found in Supplementary D.

Table 2 presents the liver segmentation results. As liver segmentation is not a challenging task, all models performed well, consistent with findings from other datasets (Bilic et al., 2023). Among the competing methods, SegResNet-based models achieved the best performance. Unlike U-Net, SegResNet employs an asymmetrical encoder-decoder architecture, where a deep encoder is optimized for feature extraction, and a compact decoder reconstructs the image (Myronenko, 2018). SegResNetVAE further enhances this architecture with a variational auto-encoder (VAE) branch at the encoder's endpoint, adding guidance and regularization (Myronenko, 2018). Such regularization is particularly beneficial in scenarios with limited training data. Given this design and its effectiveness across segmentation tasks (Myronenko, 2018), we selected SegResNetVAE as the backbone of SmoothSegNet. By employing SegResNetVAE and a global-local two-step approach, SmoothSegNet outperformed other models in liver segmentation. This approach, which optimizes liver and tumor segmentation tasks individually rather than jointly, is expected to improve performance on each subtask compared to conventional multi-class approach.

Table 3 presents the tumor segmentation results. SmoothSegNet significantly outperformed existing models, achieving a dice score of 0.547, while SegResNet, SegResNetVAE, and U-Net++ scored between 0.40 and 0.45. Although SegResNet demonstrated slightly higher sensitivity, it showed lower accuracy and dice score, indicating a tendency toward over-segmentation and increased false positives. All models maintained high accuracy and specificity, likely due to the relatively small proportion of tumor voxels compared to the total voxel count in the volume. Notably, more complex models, such as U-Net Large and MA-Net, did not outperform smaller models, suggesting that simpler architectures may generalize better in segmenting heterogeneous tumors from limited datasets. SmoothSegNet, despite having the same number of parameters as one of the larger models, achieved better performance. By using knowledge-informed label smoothing and special post-processing for tumor contours, we provided more complex models with a robust learning framework to demonstrate its advanced pattern recognition capabilities.

Table 2. Liver segmentation results across three test set replications.

| Model | Accuracy | Dice | IoU | Sensitivity | Specificity |
|---|---|---|---|---|---|
| U-Net | 0.983 (0.003) | 0.797 (0.048) | 0.672 (0.066) | 0.795 (0.066) | 0.992 (0.001) |
| U-Net Large | 0.986 (0.003) | 0.823 (0.041) | 0.709 (0.058) | 0.810 (0.048) | 0.993 (0.001) |
| U-Net++ | 0.989 (0.003) | 0.855 (0.057) | 0.759 (0.076) | 0.844 (0.096) | 0.995 (0.002) |
| MANet | 0.982 (0.004) | 0.760 (0.064) | 0.645 (0.082) | 0.749 (0.080) | 0.993 (0.001) |
| SegResNet | 0.990 (0.003) | 0.871 (0.044) | 0.780 (0.065) | 0.869 (0.067) | 0.995 (0.002) |
| SegResNetVAE | 0.990 (0.002) | 0.869 (0.034) | 0.775 (0.051) | 0.871 (0.065) | 0.995 (0.002) |
| SmoothSegNet (proposed) | **0.993 (0.003)** | **0.937 (0.019)** | **0.886 (0.034)** | **0.939 (0.023)** | **0.996 (0.001)** |

Table 3. Tumor segmentation results across three test set replications.

| Model | Accuracy | Dice | IoU | Sensitivity | Specificity |
|---|---|---|---|---|---|
| U-Net | 0.989 (0.003) | 0.325 (0.061) | 0.237 (0.055) | 0.598 (0.068) | 0.993 (0.002) |
| U-Net Large | 0.990 (0.003) | 0.341 (0.039) | 0.250 (0.039) | 0.623 (0.070) | 0.993 (0.002) |
| U-Net++ | 0.991 (0.005) | 0.407 (0.075) | 0.301 (0.060) | 0.730 (0.079) | 0.993 (0.005) |
| MANet | 0.987 (0.004) | 0.337 (0.073) | 0.247 (0.064) | 0.627 (0.105) | 0.991 (0.003) |
| SegResNet | 0.993 (0.003) | 0.454 (0.019) | 0.339 (0.022) | **0.752 (0.027)** | 0.995 (0.002) |
| SegResNetVAE | 0.992 (0.003) | 0.438 (0.031) | 0.326 (0.040) | 0.702 (0.052) | 0.995 (0.003) |
| SmoothSegNet (proposed) | **0.994 (0.004)** | **0.547 (0.069)** | **0.418 (0.064)** | 0.701 (0.112) | **0.997 (0.002)** |

### 3.3 Error Analysis

Tumor size is a major determinant of outcome after hepatic resection for HCC. Tumors larger than 10cm in diameter, called large or huge HCC, are not uncommon at clinical presentation but require specially designed surgical approaches, as nonsurgical therapies are generally considered ineffective (Y.-M. Zhou et al., 2011). Thus, we summarized model performance for tumors greater than and less than 10 cm in diameter in Table 4. As expected, all models performed reasonably well on segmenting large tumors (dice scores > 0.70) but significantly struggled segmenting smaller tumors (dice scores 0.2-0.4). SmoothSegNet achieved significantly better performance than existing methods for both large and small tumors. Notably, SmoothSegNet achieved a high sensitivity for small tumors (0.701), which are generally difficult to detect due to the low contrast background. As a baseline comparison, prior work by Morshid et al. (2019) on the same dataset achieved a dice score of 0.67 for tumors greater than 10 cm in diameter and 0.37 for those smaller than 10 cm in diameter. SmoothSegNet surpassed these benchmarks.

Table 4. Tumor segmentation results for tumors greater than and less than 10 cm in diameter.

| Tumor Size | Model | Accuracy | Dice | IoU | Sensitivity | Specificity |
|---|---|---|---|---|---|---|
| >=10cm | U-Net | 0.987 (0.004) | 0.737 (0.076) | 0.590 (0.097) | 0.691 (0.097) | 0.995 (0.001) |
| | U-Net Large | 0.987 (0.004) | 0.742 (0.081) | 0.599 (0.100) | 0.688 (0.107) | 0.993 (0.006) |
| | U-Net++ | 0.985 (0.006) | 0.739 (0.083) | 0.594 (0.106) | 0.825 (0.071) | 0.989 (0.005) |
| | MANet | 0.985 (0.008) | 0.725 (0.127) | 0.585 (0.162) | 0.743 (0.136) | 0.992 (0.007) |
| | SegResNet | 0.988 (0.005) | 0.773 (0.079) | 0.639 (0.099) | 0.816 (0.090) | 0.993 (0.002) |
| | SegResNetVAE | **0.989 (0.003)** | 0.787 (0.034) | 0.651 (0.046) | 0.790 (0.068) | 0.994 (0.002) |
| | SmoothSegNet (proposed) | **0.989 (0.004)** | **0.841 (0.022)** | **0.731 (0.028)** | **0.829 (0.060)** | **0.995 (0.001)** |
| <10cm | U-Net | 0.992 (0.005) | 0.264 (0.111) | 0.180 (0.079) | 0.551 (0.110) | 0.994 (0.004) |
| | U-Net Large | 0.991 (0.002) | 0.223 (0.028) | 0.152 (0.016) | 0.494 (0.234) | 0.993 (0.002) |
| | U-Net++ | 0.988 (0.005) | 0.240 (0.059) | 0.162 (0.041) | 0.604 (0.155) | 0.990 (0.005) |
| | MANet | 0.988 (0.003) | 0.206 (0.039) | 0.121 (0.008) | 0.443 (0.138) | 0.991 (0.002) |
| | SegResNet | 0.994 (0.003) | 0.345 (0.063) | 0.243 (0.040) | 0.666 (0.032) | 0.995 (0.004) |
| | SegResNetVAE | 0.993 (0.003) | 0.330 (0.054) | 0.231 (0.035) | 0.613 (0.064) | 0.995 (0.003) |
| | SmoothSegNet (proposed) | **0.998 (0.001)** | **0.408 (0.045)** | **0.431 (0.110)** | **0.701 (0.076)** | **0.998 (0.001)** |

Figure 4 presents segmentation examples of tumors with varying sizes and boundary characteristics, including both regular and irregular margins. Samples A and B are large tumors with a distinct dark core. Sample A has a blurry, low-contrast boundary, whereas Sample B exhibits a more well-defined margin. While all models correctly identified the dark core as tumorous, most competing methods struggled to delineate the tumor boundary. Among them, SmoothSegNet produced segmentation masks most similar to the ground truth; however, it still generated false positives and missed portions of the tumor contour in low-contrast areas (highlighted in yellow in A and B). Sample C illustrates a small tumor with a barely visible, blurry margin. Only SmoothSegNet and SegResNetVAE successfully identified the tumor location, with SegResNetVAE producing more false positives. Although SmoothSegNet's segmentation benefited from the active contour algorithm, its output remained more round-shaped compared to the actual tumor contour. Sample D represents a tumor with sparse, discontinuous components at different locations. SmoothSegNet was the only model to correctly identify both regions, yet some areas remained inaccurately segmented. Finally, Sample E showcases a tumor with a darker core and an irregular, low-contrast peripheral region.

While SmoothSegNet outperformed competing methods, it still failed to capture a small portion of the boundary (yellow box).

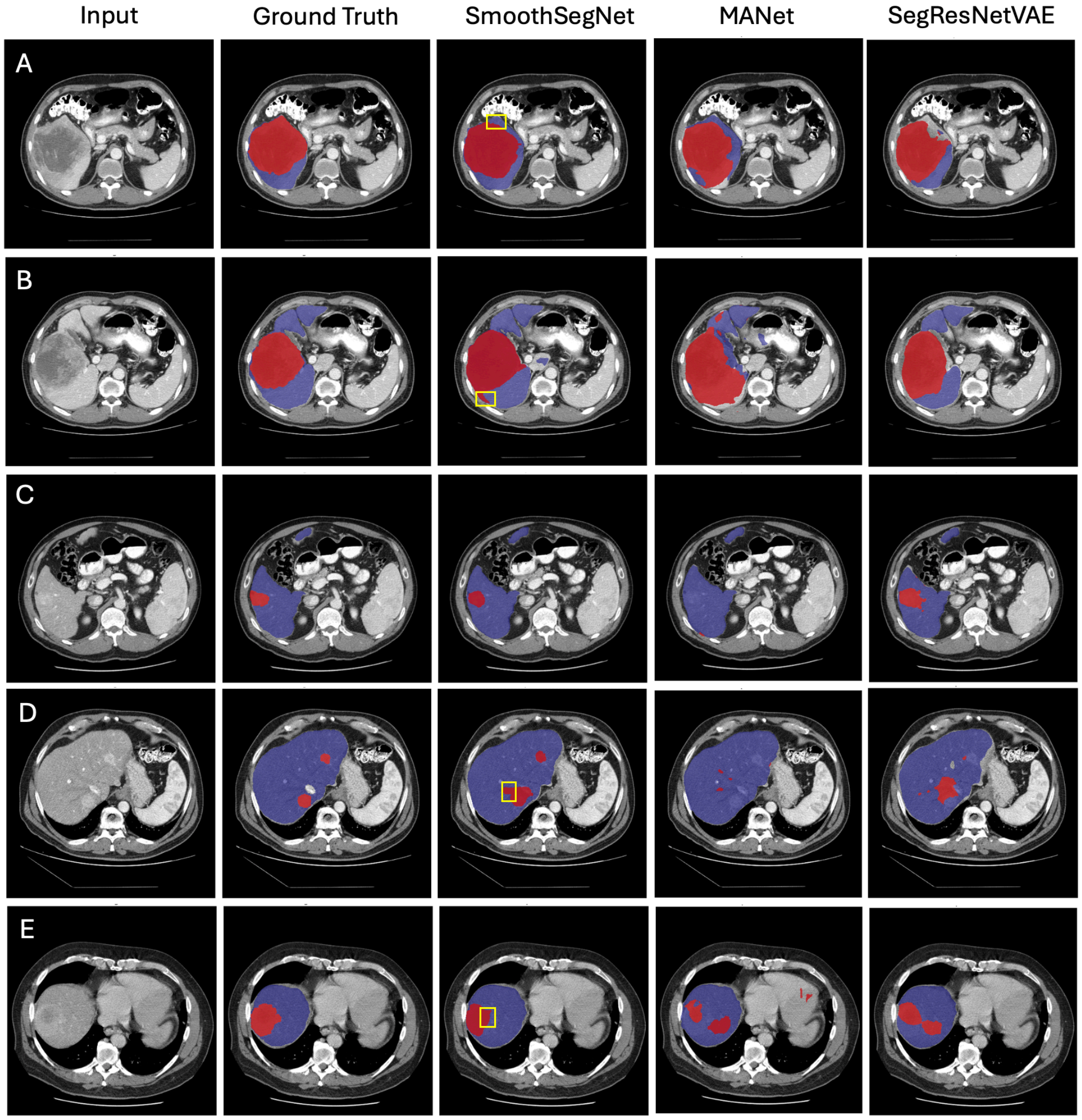

Figure 4. Sample segmentation results. A: Large tumors with blurry contour. B: Large tumor with well-defined contour. C: Small tumor with blurry and irregular contour. D: Small tumor with sparse pieces at different locations. E: Small tumor with irregular margin. Yellow box: failure cases of SmoothSegNet.

Segmentation examples of a representative set of large and small tumors with regular and irregular boundaries are shown in Figure 4. Samples A and B are two large tumors with a visible dark area as the tumor core. The former has a blurry, light-contrast boundary and the latter has a more well-defined boundary. All models were able to detect the dark area as tumorous, however, most competing methods failed at delineating the boundary. SmoothSegNet generated segmentation mask was the most similar to the ground truth, but it still produced some false positives and missed some tumor contour in the low contrast areas (yellow circle in A and B). Sample C is an example of a small tumor that is hardly visible with blurry margin. SmoothSegNet and SegResNetVAE were the only two models able to correctly locate the tumor location. SegResNetVAE generated more false positives. While SmoothSegNet's segmented shape was close to the ground truth with the help of the active contour algorithm, its segmented shape is still a quite round-shaped compared to the ground truth tumor contour. Sample D is an example of a tumor with sparse pieces located at different locations. SmoothSegNet was the only model correctly locating the two pieces, but the model outputs still have areas of improvement. Sample E is an example of a small tumor with a darker core area and surrounding tumor with irregular margins and low contrast. While SmoothSegNet performed the best among the competing methods, it still missed a small portion of the boundary (highlighted in yellow box).

### 3.4 Ablation Study

To better understand the contribution of the three key components of the proposed model, we conducted an ablation study to examine the impact of (i) the two-step global-local segmentation pipeline, (ii) knowledge-informed label smoothing, and (iii) active contour-based boundary refinement. The ablation study results are shown in Table 5.

For liver segmentation, the multi-class approach resulted in poorer liver segmentation performance. This is expected, as simultaneously learning liver and tumor segmentation forces the model to balance its performance on both tasks. In contrast, the two-step approach yielded better liver segmentation performance, as it allowed the model to focus solely on the liver segmentation task without the added complexity of tumor segmentation. There is no significant improvement gained by applying contour adjustment since the liver has a regular boundary, hence we did not include the results with contour adjustment. Similarly, label smoothing is not applicable in liver segmentation.

For tumor segmentation, incorporating knowledge-informed label smoothing (Model III) improved segmentation performance, highlighting the benefits of this regularization technique. Additionally, applying active contour-based boundary refinement further enhanced segmentation, particularly in terms of the sensitivity metric, recovering detection capabilities that were compromised in the multi-class model. This suggests that the post-processing step was valuable for refining segmentation outputs and ensuring the accurate detection of all tumors.

By integrating all three components, the proposed SmoothSegNet achieved the best overall performance for both liver and tumor segmentation. These results confirm that these components work synergistically, contributing to state-of-the-art segmentation performance for both liver and tumors.

Table 5. Ablation study of the proposed framework in terms of liver and tumor segmentation performance. Smoothing: clinical knowledge-informed label smoothing; Contour adjust: active contour algorithm for tumor boundary refinement.

| Model version | Two-step | Contour adjust | Smoothing | Dice | Sensitivity |
|---|---|---|---|---|---|
| Liver segmentation | | | | | |
| I | O | NA | NA | 0.894 | 0.900 |
| V (proposed) | X | NA | NA | 0.942 | 0.955 |
| Tumor Segmentation | | | | | |
| I | O | O | O | 0.498 | 0.702 |
| II | X | O | O | 0.472 | 0.513 |
| III | X | O | X | 0.500 | 0.589 |
| IV | X | X | O | 0.563 | 0.741 |
| V (proposed) | X | X | X | 0.570 | 0.801 |

## 4. Conclusion

This research introduces SmoothSegNet, a deep learning-based framework for liver and tumor segmentation with knowledge-informed label smoothing. Each of the key components of this framework are designed to tackle challenges of tumor segmentation, addressing model overfitting, and achieving more accurate delineation of tumor boundaries. Specifically, SmoothSegNet distills knowledge from clinical data and uses knowledge-informed smooth label to regularize training and mitigate overfitting. SmoothSegNet is also distinctive from conventional multi-class approaches in its two-step global and local view segmentation process, which divide the original challenging task into two subtasks, enabling task-specific optimization across the preprocessing, training, and post-processing phases. To tackle the imprecise margin of some tumors, we incorporated an active contour algorithm to refine segmented tumor contours. These three design components synergistically contribute to the overall framework achieving state-of-the-art segmentation performance for liver tumors.

This study has several limitations. One limitation of this study is the constrained sample size of labeled data. While the proposed model is designed to mitigate this issue by employing knowledge-informed label smoothing, the importance of a sufficient number of labeled samples for training an accurate and robust model cannot be overstated. Expanding the number of labeled samples not only has the potential to enhance segmentation accuracy for the current backbone models but also enables training of deeper backbone models, which may further improve the current segmentation performance. Additionally, tumor size is another critical factor influencing model training. Future work could involve developing specialized models for large and small tumors to further enhance segmentation accuracy. For small tumors, a larger dataset and more precise training guidance are essential, including details on the number of lesion areas, tumor sizes, and boundaries. Furthermore, incorporating a more comprehensive set of clinical features such as comorbidities and treatment history could provide higher quality knowledge guidance for model training. Although liver tumor segmentation demands further research to improve segmentation accuracy, this work highlight the potential of these AI tools to assist with tumor segmentation.

## Appendix

### Appendix A. Feature Selection for TLVR

We aimed to identify the clinical features most predictive of the TLVR. According to feature contributions in linear regression, we selected the top $n$ features most predictive for $R_i$ using 5-fold cross-validation (CV). The optimal number of features, $n$, was determined to be 15, based on achieving the highest Pearson Correlation coefficient between $R$ and $\hat{R}$ across the 5-fold CV. Table B1 shows selected top 15 most predictive features in descending order of importance, along with their descriptions.

Table A1. Clinical variables included for knowledge extraction.

| *Clinical Feature* | *Description* |
|---|---|
| *T_involvement* | Liver involvement by the tumor, either less than 50% involvement of liver or more. |
| *Personal history of cancer* | Whether the patient has cancer before. |
| *Lymphnodes* | Whether the patient has lymphnodes. |
| *CLIP_Score* | CLIP score. |
| *TNM* | TNM staging. |
| *Metastasis* | Whether cancer cells spread from the original tumor site to distant parts of the body, forming new tumors in organs or tissues. |
| *Evidence_of_cirh* | Whether there is evidence of late stage of scarring (fibrosis) of the liver. |
| *Alcohol* | Whether the patient drinks alcohol or not. |
| *AFP* | Alpha fetoprotein level (ng/ml) obtained from blood test. |
| *Smoking* | Whether the patient smokes or not. |
| *Diabetes* | Whether the patient has diabetes. |
| *fhx_can* | Whether the patient's family has history of cancer. |
| *age* | The age of patient. |
| *TTP* | Number of weeks between the procedure and the first evidence of tumor progression. |
| *Interval_BL* | Number of Days between HCC diagnosis (either by previous imaging or biopsy) and pre- procedural (baseline) CT. |

**Appendix B. Model Training and Validation Procedure**

The training process for SmoothSegNet consists of two sequential steps. First, a liver segmentation model is trained taking entire CT scans as input. Then, a tumor segmentation model is trained on cropped liver regions delineated based on ground truth liver masks. Each segmentation stage involves its own pre- and post-processing steps, as previously described. Note that SmoothSegNet is model agnostic, allowing researchers to choose the backbone model that best suits their dataset or to adapt state-of-the-art architectures accordingly. During evaluation, each test sample follows the same two-step segmentation process, where predicted liver masks are input for tumor segmentation. No clinical features are required at inference, making the model more practical and versatile for deployment. The hyperparameter settings are detailed in Supplementary B.

**Appendix C. Task-specific Pre- and Post-processing for Liver and Tumor Segmentation**

Proper preprocessing and post-processing steps can greatly enhance segmentation results. Preprocessing addresses imaging heterogeneity by standardizing inputs and enhancing the visibility of relevant anatomical structures. Following segmentation, post-processing techniques refine the outputs by improving coherence in the liver class and accurately delineating complex tumor boundaries. In this section, we aim to provide optimal preprocessing and post-processing methods for liver tumor segmentation based on our empirical results (Figure C1), with the goal of offering guidance for future related studies.

The parameters used in the pre-and post-processing steps are provided in Supplementary B. Thanks to the sequential framework design, the model provides great flexibility in leveraging various pre- and post-processing steps. These steps can vary on a case-by-case basis, and other researchers can choose different processing steps for their own segmentation tasks or datasets. Details of the task-specific pre- and post-processing for liver and tumor segmentation are provided in Supplementary C.

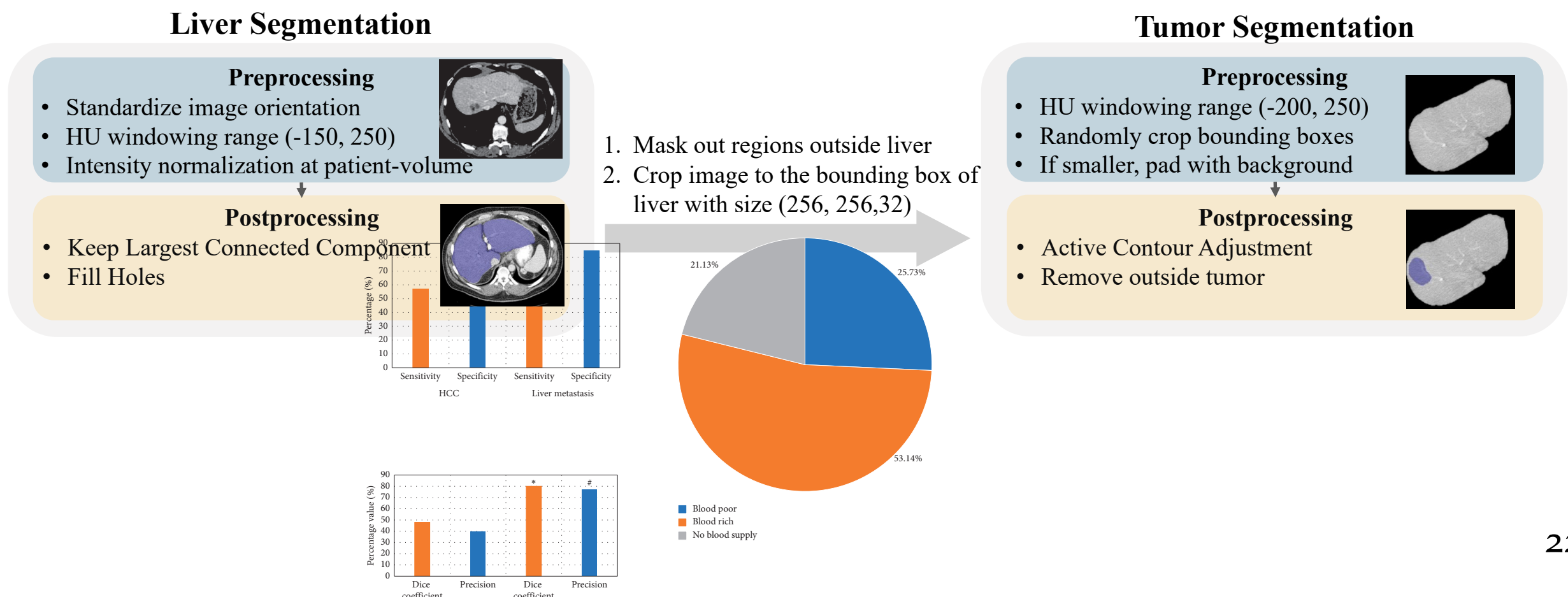

Figure C1. Task-specific pre- and post-processing for liver and tumor segmentation with representative examples demonstrating the output at each pre- and post-processing step.

**Appendix D. Knowledge-informed Label Smoothing Reduced Overfitting**

To further understand the effect of the proposed label smoothing in model training, we conducted a comparative analysis of the training loss and validation dice scores using the same model under identical settings. We compared two versions of SmoothSegNet with and without using knowledge-informed label smoothing in Figure D1. The model incorporating label smoothing exhibited a more gradual decrease in training loss and a more rapid increase in the validation dice score, whereas the model without label smoothing's training loss quickly converged and its validation metric saturated early. Thus, label smoothing effectively reduced overfitting and improved the model's generalizability to unseen data.

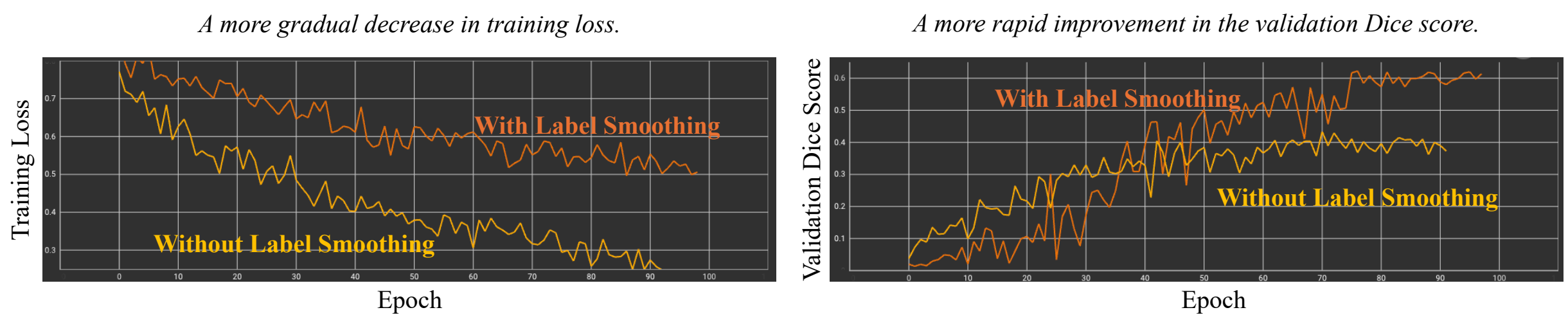

Figure D1. Training and validation curves with vs. without knowledge-informed label smoothing. Label smoothing yields a more gradual decrease in training loss and a more rapid improvement in the validation metric, reducing overfitting during model training.

**Funding.** No specific funding was received for this work.

**Role of the Funder.** Not applicable.

**Disclosure of Interest.** The authors report no conflict of interest.

**Consent and Approval statement.** Institutional review board (IRB) approval was not required for this study, as it only used publicly available data.

# Supplemental Online Material

## Supplementary A. Overview and Preparation of the HCC-TACE-Seg Dataset

We utilized the HCC-TACE-Seg dataset from TCIA, which contains retrospectively collected pre- and post-procedural CT images of 105 confirmed HCC patients who underwent TACE between 2002 and 2012 (Moawad et al., 2021). After excluding patients with missing or misaligned segmentation files, we extracted pre-procedural CT scans of liver tumors from 98 patients, along with corresponding segmentation labels for five classes of objects: background, liver, tumor mass, portal vein, and abdominal aorta (Figure SA1). Segmentations were created using AMIRA (Stalling et al., 2005) and manually curated by clinicians. To focus on the liver and tumor classes, we masked the portal vein as liver and the abdominal aorta as background. Each CT scan has spatial dimensions 512 x 512 x N, where the number of slices N varies by patient. Additionally, a clinical data file containing related demographics, medical history, diagnosis, and treatment response is provided in HCC-TACE-Seg. From these clinical variables, we selected variables that are known risk factors of liver cancer (Mohammadian et al., 2018; Morshid et al., 2019; Shah et al., 2023). The dataset was randomly split into 80% training and 20% test sets, with the train-test split repeated for three replications.

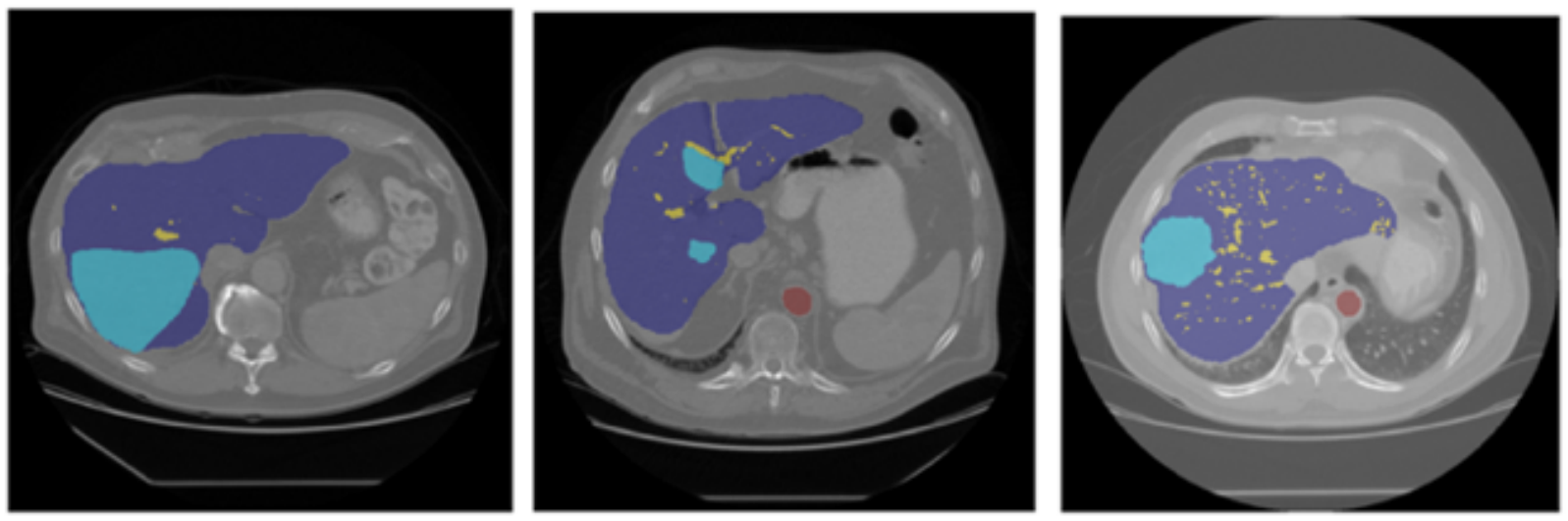

Figure SA1. Sample CT scan slices from the HCC-TACE-Seg dataset with overlaid segmentation labels. Purple: liver; blue: tumor; yellow: portal vein; red: abdominal aorta.

## Supplementary B. Hyperparameter Settings of SmoothSegNet

Table SB1. Liver segmentation model parameters

| Parameters | Description | Value |
| --- | --- | --- |
| input_image_size | the size of images to input into the network | (512, 512, 16) |
| vae_default_std | if not to estimate the std, use the default value | 0.3 |
| vae_nz | number of latent variables in VAE | 256 |
| blocks_down | number of down sample blocks in each layer | [1, 2, 2, 4] |
| blocks_up | number of up sample blocks in each layer | [1, 1, 1] |
| init_filters | number of output channels for initial convolution layer | 16 |
| in_channels | number of input channels for the network | 1 |
| out_channels | number of output channels for the network | 2 |
| hu_range | an interval of intensity values on the Hounsfield scale to enhance the contrast of target tissues | (-150, 250) |

Table SB2. Tumor segmentation mode parameters

| Parameters | Description | Value |
| --- | --- | --- |
| input_image_size | the size of images to input into the network | (256, 256, 32) |
| vae_nz | number of latent variables in VAE | 256 |
| blocks_down | number of down sample blocks in each layer | [1, 2, 2, 4] |
| blocks_up | number of up sample blocks in each layer | [1, 1, 1] |
| init_filters | number of output channels for initial convolution layer | 16 |
| in_channels | number of input channels for the network | 1 |
| out_channels | number of output channels for the network | 3 |
| hu_range | an interval of intensity values on the Hounsfield scale to enhance the contrast of target tissues | (-200, 250) |
| lambda_weak | weight of knowledge-informed soft labels | 0.5 or 0.6 |

Table SB3. Training setting parameters

| Parameters | Description | Value |
|---|---|---|
| max_epochs | max number of epochs in training | 150 |
| batch_size | batch size | 1 |
| learning_rate | learning rate | 1e-4 |
| optimizer | algorithm used for optimizing the neural network weights | Adam |
| weight_decay | weight decay (L2 penalty) of the optimizer | 1e-5 |
| lr_scheduler | strategy to adjust the learning rate during training | CosineAnnealingLR |

## Supplementary C. Task-specific Pre- and Post-processing for Liver and Tumor Segmentation

### SC.1 Image Preprocessing

Image preprocessing is a critical step to reduce imaging differences across CT scans and ensure that the deep learning models can learn effectively. We designed preprocessing steps to optimize the input data for both liver and tumor segmentation, addressing challenges such as image heterogeneity and enhancing the focus on relevant anatomical structures. The preprocessing pipeline for the liver segmentation model begins with loading raw medical imaging data and standardizing the image orientation. Hounsfield units (HU) windowing was applied to enhance the visibility of liver and tumor by adjusting intensity values to ranges recommended in the literature (Kim et al., 2020). Intensity normalization is then performed across the dataset. For the tumor segmentation model, additional preprocessing steps were implemented after standardizing the image orientation. Assuming that liver tumors are located inside the liver (Tummala et al., 2017), the inputs to the tumor segmentation model are simplified by masking out regions outside the liver as background. This approach narrows the model's focus to distinguishing healthy and tumorous tissues within the liver, eliminating the need to navigate uninformative and potentially distractive features from surrounding structures. Given that each patient's liver varies in size, the bounding box containing the liver is randomly cropped to a fixed size with padding as needed.

### SC.2 Liver and Tumor Segmentation Post-processing

In the postprocessing phase of the proposed segmentation pipeline, the segmentation results for both the liver and tumor classes are refined, with a specific focus on enhancing the coherence of the liver segmentation while cautiously handling tumor regions due to their complex characteristics. For the liver class, a strategy to retain only the largest connected components and then apply a fill holes technique was employed. This approach is motivated by the anatomical consistency of the liver, which typically forms a single, large volume within the abdominal cavity. By keeping the largest connected component, we remove any small, spurious regions that were incorrectly labeled as liver tissue during the segmentation process. Filling holes within this largest component helps to ensure that the liver's representation is solid and continuous, correcting for any internal inaccuracies that might have arisen during segmentation.

The post-processing assumptions for liver segmentation does not apply to the case of tumor. Given the inherent variability and complexity of tumor appearances, tumors can exhibit a wide range of sizes, shapes, and may sometimes be composed of multiple disjointed parts, especially in cases of metastatic or multifocal liver disease. Applying the keep largest component and fill holes strategy might inadvertently eliminate smaller, yet clinically significant tumor regions. Instead, we employed an active contours algorithm for detecting contour without borders for refining tumor segmentation outputs (Chan & Vese, 2001; Márquez-Neila et al., 2014). In classical active contour algorithms, an edge-based energy function determines when to stop the curve evolution based on gradient of the image. However, these algorithms suffer when the object lacks clear edges or when the image is noisy. Thus, we employed an active contour algorithm that is capable of detecting smooth boundaries without relying on image gradients (Chan & Vese, 2001). This algorithm iteratively identifies the boundary curve that differentiates energy inside and outside the object. We provided the tumor segmentation outputs as the initial curve and run the algorithm for two iterations to refine the boundary. This method is adept at capturing the irregular tumor boundaries in noisy CT images, which is crucial for achieving high fidelity in the segmentation of complex tumor shapes.

## Supplementary D. Benchmark Methods and Evaluation Metrics

### SD.1 Benchmark Methods and Training Details

We compared SmoothSegNet's performance against various state-of-the-art benchmark methods for medical image segmentation on this dataset. The competing models included:

- U-Net (Kerfoot et al., 2019): an improved version of U-Net with residual units. We trained two versions of the U-Net model: a four-layer U-Net with 32, 64, 128, 256 convolutional kernels in each up-sampling layer and two residual units, and a five-layer U-Net with the fifth layer having 512 convolutional kernels, same as the original paper (Kerfoot et al., 2019), denoted as U-Net Large.
- U-Net++ (Zhou et al., 2018): an improved U-Net with nested skip pathways between encoder and decoder blocks at multiple resolutions.
- SegResNet and SegResNetVAE (Myronenko, 2018): an asymmetric network featured by a large encoder and a compact decoder, and an optional variational auto-encoder (VAE) to regularize reconstruction. We trained two versions of this network, with and without the VAE.
- MA-Net (Fan et al., 2020): a variant of U-Net with multi-scale attention network;

We applied instance normalization and a dropout probability of 0.2 across all models, utilizing these methods for multi-class segmentation to simultaneously segment both the liver and tumor. All models were implemented in PyTorch and trained on a NVIDIA L4 16GB GPU using Adam optimizer with learning rate 1e-4. Parameter settings details are provided in Supplementary B.

### SD.2 Evaluation Metrics

We evaluated segmentation performance class-wise for liver and tumor using the following metrics:

- **Accuracy** measures the proportion of correctly segmented pixels within the all pixels in each 3D CT image. It is calculated as $\frac{TP+TN}{TP+TN+FP+FN}$, where $TP$, $TN$, $FP$, and $FN$ represent the numbers of true positives, true negatives, false positives, and false negatives, respectively.
- **Dice score** is a statistic commonly used in image segmentation to gauge the similarity of two masks. The dice score can be defined as $\frac{2\times TP}{2\times TP+FP+FN}$. This metric highlights the importance of true positives by considering the size of the intersection over the average size of two samples.

- **Jaccard Index**, also known as **Intersection over Union (IoU)**, is commonly used metric in image segmentation to measure the similarity between the predicted and ground truth masks. It is defined as the ratio of the intersection of the predicted and actual segmentation regions to their union, given by $\frac{TP}{TP+FP+FN}$.
- **Sensitivity** measures the proportion of actual positives that are correctly identified, calculated as $\frac{TP}{TP+FN}$. This metric indicates the model's ability to correctly detect positives out of all actual positive cases, which is particularly important in medical diagnosis where missing a positive case can have serious consequences.
- **Specificity** measures the proportion of correctly identified negative pixels (true negatives) out of all actual negative pixels in image segmentation, defined as $\frac{TN}{TN+FP}$. It quantifies the ability to avoid false positives.